%% file: emnlp2020.tex
\documentclass[11pt,a4paper]{article}
\usepackage[hyperref]{emnlp2020}
\usepackage{times}
\usepackage{latexsym}

\usepackage{color,colortbl}
\usepackage{xspace}
\usepackage{makecell}

\usepackage{musicography}
\usepackage{figsize}
\usepackage[normalem]{ulem}
\usepackage{xspace}
\usepackage{amsmath}
\usepackage{amssymb}
\usepackage{arydshln}
\usepackage{enumitem}
\usepackage{url}
\usepackage{soul}
\usepackage{xcolor}
\usepackage[ruled,noline,linesnumbered]{algorithm2e}
\usepackage{pifont}% http://ctan.org/pkg/pifont
\usepackage{graphicx}
\usepackage{pgfplots}
\pgfplotsset{compat=1.16} 
\usepackage{enumitem}
\usepackage{makecell} 
\usepackage{xspace}
\usepackage{comment}
\usepackage{ragged2e}
\usepackage{booktabs,amsfonts,dcolumn}
\newcolumntype{d}[1]{D..{#1}}
\usepackage{color, colortbl}
\usepackage{capt-of}

\usepackage{balance}

\newcommand{\ourmodel}{{{\textsc{Delorean}}}\xspace}
\newcommand{\modelname}{{{\textsc{Delorean}}}\xspace}
\newcommand{\counterdata}{{{\textsc{TimeTravel}}}\xspace}
\newcommand{\dataset}{{\usefont{T1}{pzc}{m}{n} ART}}
\newcommand{\comet}{COMET}
\newcommand{\zs}{Zero-Shot}
\newcommand\lmallcometembprefix{\textit{+\comet-Emb}}

\usepackage{tikz}
\newcommand{\fwvector}{\includegraphics[scale=0.35]{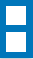}}%
\newcommand{\bwvector}{\includegraphics[scale=0.35]{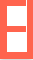}}%
\newcommand{\mixvector}{\includegraphics[scale=0.35]{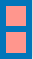}}%

\usepackage{microtype}

\aclfinalcopy % Uncomment this line for the final submission
\usepackage{amsmath,amsfonts,amssymb,bbm,epsfig,bm}
\usepackage{array}
\usepackage{booktabs}

\usepackage{algorithm2e}
\usepackage{algorithmic}

\usepackage{color}

\newcommand{\LM}{\text{LM}}

\newcommand{\dataName}{{{\textsc{TimeTravel}}}\xspace}

\title{Back to the Future: Unsupervised Backprop-based Decoding for\\ Counterfactual and Abductive Commonsense Reasoning}
\author{Lianhui Qin\textsuperscript{$\dagger\ddagger$} \hspace{.3cm}  Vered Shwartz \textsuperscript{$\dagger\ddagger$} \hspace{.3cm} Peter West \textsuperscript{$\dagger\ddagger$} \hspace{.3cm} Chandra Bhagavatula\textsuperscript{$\ddagger$} \\ \textbf{ \hspace{.3cm} Jena D. Hwang \textsuperscript{$\ddagger$}  \hspace{.3cm} Ronan Le Bras \textsuperscript{$\ddagger$} \hspace{.3cm}  Antoine Bosselut \textsuperscript{\musNatural$\ddagger$}  \hspace{.3cm} Yejin Choi\textsuperscript{$\dagger\ddagger$} }\\
\textsuperscript{$\dagger$}Paul G. Allen School of Computer Science \& Engineering, University of Washington\\
\textsuperscript{$\ddagger$}Allen Institute for Artificial Intelligence \hspace{.3cm} 
\textsuperscript{\musNatural}Stanford University  \\
\texttt{\{lianhuiq, pawest, yejin\}@cs.washington.edu} \\
\texttt{\{vered, chandrab, jenah, ronanlb\}@allenai.org}
}

\date{}

\begin{document}

\maketitle

\begin{abstract}
Abductive and counterfactual reasoning, core abilities of everyday human cognition, 
%involve hypothesizing with respect to past and future observations. 
require 
%generating natural language hypotheses on 
reasoning about 
what might have happened at time $t$, while conditioning on multiple contexts from the relative past and future.
However, simultaneous incorporation of past and future contexts using generative language models (LMs) can be challenging, as they are trained either to condition only on the past context or to perform narrowly scoped text-infilling. 

In this paper, we propose \textsc{DeLorean}, a new unsupervised decoding algorithm that can flexibly incorporate both the past and future contexts using only off-the-shelf, left-to-right language models and no supervision. The key intuition of our algorithm is incorporating the future through \emph{back-propagation}, during which, we only update the internal representation of the output while fixing the model parameters. By alternating between forward and backward propagation, \textsc{DeLorean} can decode the output representation that reflects both the left and right contexts. 
%
%We propose an unsupervised LM-based approach that consists of alternating passes: a forward pass that concerns maintaining coherence with the past, and a backward pass that incorporates task-specific future constraints through backpropagation. 
%
%We demonstrate our approach on the abductive and counterfactual generation tasks, where it significantly improves over a range of other unsupervised and some supervised methods, in terms of both automatic and human evaluation. %
We demonstrate that our approach is general and applicable to two nonmonotonic reasoning tasks: abductive text  generation and counterfactual story revision, where \textsc{DeLorean} outperforms a range of unsupervised and some supervised methods, based on automatic and human evaluation.\footnote{Code is available at \url{https://github.com/qkaren/unsup_gen_for_cms_reasoning}}

% Abductive and counterfactual reasoning, as common activities of human cognition, involves inferring plausible explanations of observed consequences or predicting alternative future events given changed conditions. Such nonmonotonic reasoning is particularly challenging as a model is required to consider both the past and future contexts in order to generate a hypothesis. This contrasts with the recent powerful pre-trained language models (LMs), such as GPT2, that facilitate only left-to-right generation. This paper develops an unsupervised decoding strategy for pre-trained LMs to perform nonmonotonic reasoning. Our approach consists of alternating forward and backward passes, where the backward pass brings in any task-specific future information into the token decoding distributions and the forward pass refines the distributions fluency and coherence with the past. We further develop an unsupervised ranking approach that selects the best hypothesis from generated candidates. Human and automatic evaluations show our approach significantly improves over a range of other unsupervised and some supervised methods on the two reasoning tasks. 

% Nonmonotonic reasoning is the critical ability of humans to make plausible, but fallible inferences for counterfactual events, or incomplete observations. 
\end{abstract}

\input{1-intro}

\input{2-background}

\input{3-approach}

\input{4-abductive}
\input{5-counterfactual}
\input{6-related}

\input{7-conclusion}

\balance
\bibliography{anthology,emnlp2020}
\bibliographystyle{acl_natbib}

\include{appendix}

\end{document}

%% file: 1-intro.tex
\begin{figure}[t!]
\centering
\includegraphics[width=\linewidth]{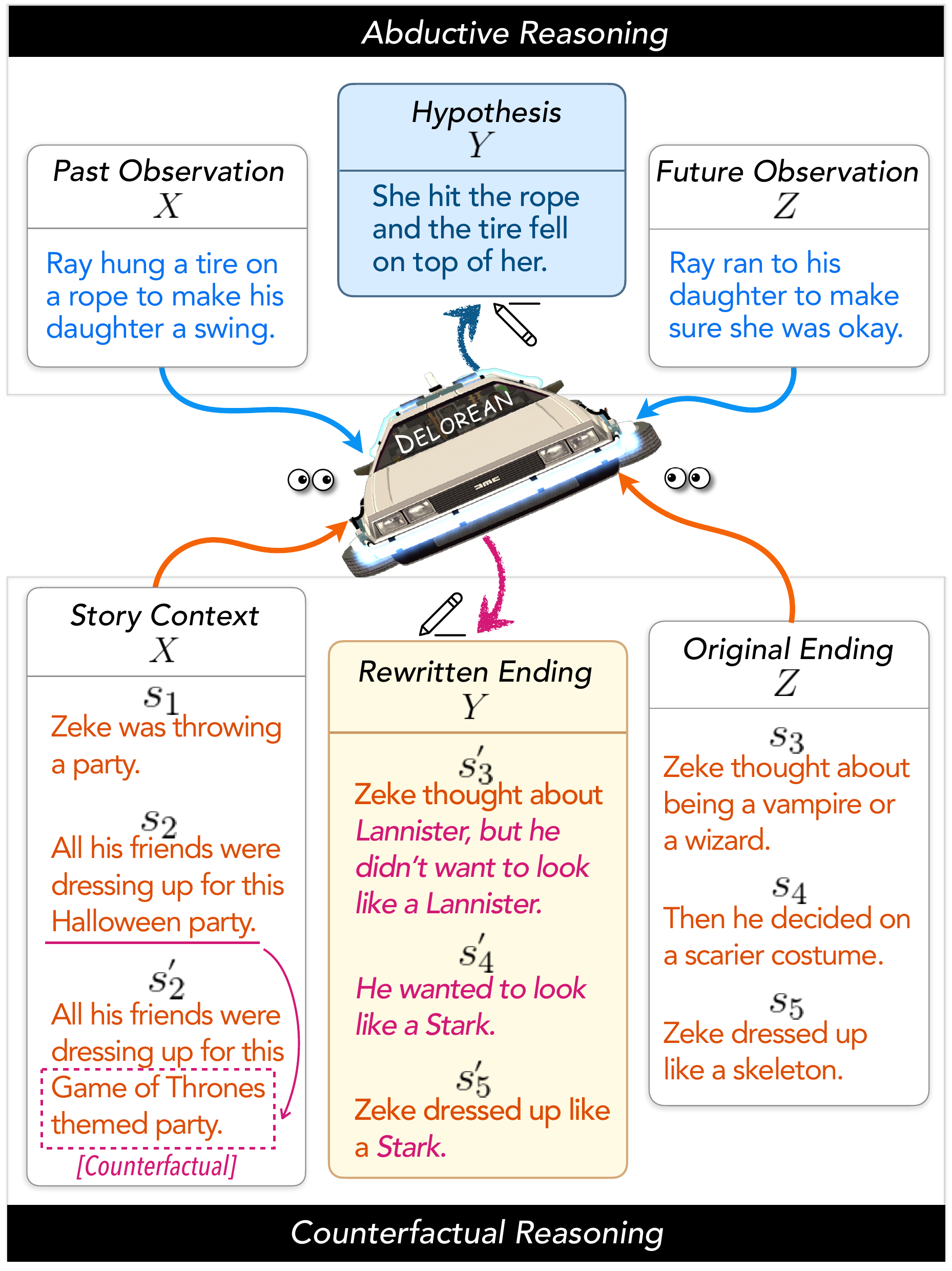}
\vspace{-10pt}
% \caption{Our proposed method, \modelname, with predictions generated for abductive (top) and counterfactual (bottom) reasoning. In abductive reasoning, a left context $X$ and a right context $Z$ are observed, and the goal is to generate a hypothesis $Y$ that explains the incomplete observations. In counterfactual reasoning, a story beginning $X$, an original story ending $Z$, and a counterfactual condition that breaks the original ending are given. The goal is to generate an edited ending $Y$ that makes minimal edits to $Z$, but achieves coherence with the counterfactual situation. Our method alternates forward (left-to-right) and backward (right-to-left) passes that iteratively integrate context information from both sides and refine the generated texts.
% }
\caption{\modelname, our proposed method, with generated reasoning results. \textbf{Top}: the goal in abductive reasoning is to generate a hypothesis ($Y$) of what happened between the observed past ($X$) and future ($Z$) contexts. \textbf{Bottom}: In counterfactual reasoning, given a story context altered by a counterfactual condition, $X$, and the original ending $Z$, the goal is to generate a new ending $Y$ which is coherent with $X$ while remaining similar to $Z$. The story from \dataName~\citep{qin2019counterfactual} consists of five sentences. Our approach alternates forward (left-to-right) and backward (right-to-left) passes that iteratively refine the generated texts w.r.t context from each side.
}
\vspace{-10pt}
\label{fig:main_example} 
\end{figure}

\section{Introduction}
\label{sec:intro}

Everyday causal reasoning requires reasoning about the likely explanations to partially observable past and future (\emph{abductive} reasoning \cite{peirce1960collected}) and reasoning about the alternative future based on counterfactual past (\emph{counterfactual} reasoning). Such \emph{nonmonotonic} reasoning requires inferring plausible but potentially defeasible conclusions from incomplete or hypothetical observations \cite{reiter1988nonmonotonic}. While humans are remarkably good at this type of causal reasoning, developing AI systems capable of nonmonotonic reasoning for a wide range of situations describable in natural language has been a major open research question. 
%developing machines with causal reasoning abilities is a fundamental challenge.

%In this paper we target two tasks involving nonmonotonic reasoning: \emph{abductive reasoning} and \emph{counterfactual reasoning}. 
More concretely, with abductive reasoning, the goal is to find the most plausible explanation for incomplete observations \cite{peirce1960collected}. In the top part of Figure~\ref{fig:main_example}, given the first observation that Ray is ``making his daughter a swing" and the later observation that he ``ran to [her] to make sure she was okay," we can hypothesize that she somehow got hurt by the swing.  

In contrast, counterfactual reasoning concerns the causal changes to future events given a change in the past condition \cite[i.e., ``counterfactual condition'';][]{goodman1947problem}. For example, the bottom part of Figure~\ref{fig:main_example} shows the \emph{original} five sentence story ($S_1$, ..., $S_5$) and an alternative \emph{counterfactual condition} given in $S_2'$---that instead of being a generic ``Halloween party'', the new counterfactual condition is that it is going to be a ``Game of Thrones themed party''! Given these, the problem we want to solve is to update the future events ($S_3', ..., S_5'$), so that instead of ``Zeke dressed up as skeleton'', we have ``Zeke dressed up like a Stark''.\footnote{``Lannister'' in $S_3'$ and ``Stark'' in $S_4'$ and $S_5'$ refer to character names in the TV show, ``Game of the Thrones.'' All the output text shown in Figure~\ref{fig:main_example} is the actual system output from \textsc{DeLorean}.}

Recently, two tasks and corresponding benchmarks have been introduced to tackle language-based nonmonotonic reasoning: the \dataset{} dataset for abductive NLG \cite{bhagavatula2019abductive}, and the \dataName~dataset for counterfactual story rewriting \cite{qin2019counterfactual}. 
Both tasks are framed as conditional generation, with multiple contexts to condition on. 
%in which the goal is to generate a natural language output that follows a given observation and adheres to certain reasoning constraints. 
%The large-scale datasets that were collected for these tasks, the \dataset{} dataset for abductive reasoning and the \dataName~dataset for counterfactual reasoning, have encouraged researchers to develop supervised learning approaches to these tasks, usually by fine-tuning large pre-trained language models (LMs), such as GPT2 \cite{gpt2}. 
The currently dominant paradigm for conditional text generation tasks is fine-tuning pre-trained language models (LMs), such as GPT2 \cite{gpt2}, on large-scale  training data for supervision. 
%Due to the rich knowledge captured during the pre-training phase, such approaches have significantly improved performance on these benchmarks. 
However, despite the large number of training examples, supervised approaches still perform considerably worse than humans and are subject to developing superficial strategies such as repeating the observations as is or memorizing prevalent surface patters specific in the dataset \cite{qin2019counterfactual}. Furthermore, having to require large-scale training data for each domain and task would be utterly inefficient for broad-coverage nonmonotonic reasoning in language.

In this paper, we investigate an alternative path toward language-based nonmonotonic reasoning using pre-trained language models as is. Intuitively, both the abductive and counterfactual reasoning requires learning coherent patterns in narrative, which should be already available in large-scale pretrained language models. However, the key challenge is that most generative language models are trained to condition only on the left context, or to perform narrowly scoped text-infilling.

%However, approaching nonmonotonic tasks in an unsupervised manner is not trivial either. Most available pre-trained LMs facilitate only left-to-right prediction (i.e., conditioning on a left context), in contrast to the ``in-fill'' nature of the nonmonotonic tasks, which requires simultaneously incorporating both the past and future contexts.

This paper presents \modelname: DEcoding for nonmonotonic LOgical REAsoNing, an unsupervised decoding algorithm that only assumes off-the-shelf left-to-right language models with no supervision. The key intuition of our algorithm is incorporating the future through back-propagation, during which, we only update the internal representation of the output while fixing the model parameters.
%Our unsupervised LM decoding strategy consists of alternating forward and backward passes that each refine the generated sequence with respect to a different objective. 
More specifically, \modelname alternates between the forward and backward passes, where the forward pass performs left-to-right inference given the left context (roughly maximizing $P({Y}|X)$ in Figure~\ref{fig:main_example}), while the backward pass instills the right constraint through right-to-left backpropagation with a task-specific loss (roughly maximizing $P(Z|{XY})$). The forward and backward outputs are mixed into a single vector, from which tokens are sampled to generate the desired output. To choose the best output across iterations, we employ an unsupervised ranking step based on BERT's next sentence prediction task to measure coherence \cite{devlin2018bert}.
% We further develop an unsupervised ranking method that uses BERT~\cite{devlin2018bert} next sentence prediction for measuring coherence.

% We conduct both human and automatic evaluations on the two tasks. The proposed approach substantially improves over other unsupervised methods, and performs competitively or better than some supervised ones. 

On both tasks, \modelname outperforms all other unsupervised methods in terms of both automatic metrics and human evaluation, demonstrating that nonmonotonic reasoning through conditional decoding is a promising research direction. Moreover, outputs produced by our model are judged as more coherent than those from the supervised models. In sum, our study shows that backpropagation-based decoding may enable additional future applications of unsupervised generation and reasoning.
%Our results demonstrate that nonmonotonic reasoning through generation is a promising research direction and that backpropagation-based decoding can enable new applications of language models for unsupervised generation and reasoning applications. 

\begin{figure*}[t]
\centering
{\includegraphics[scale=0.48]{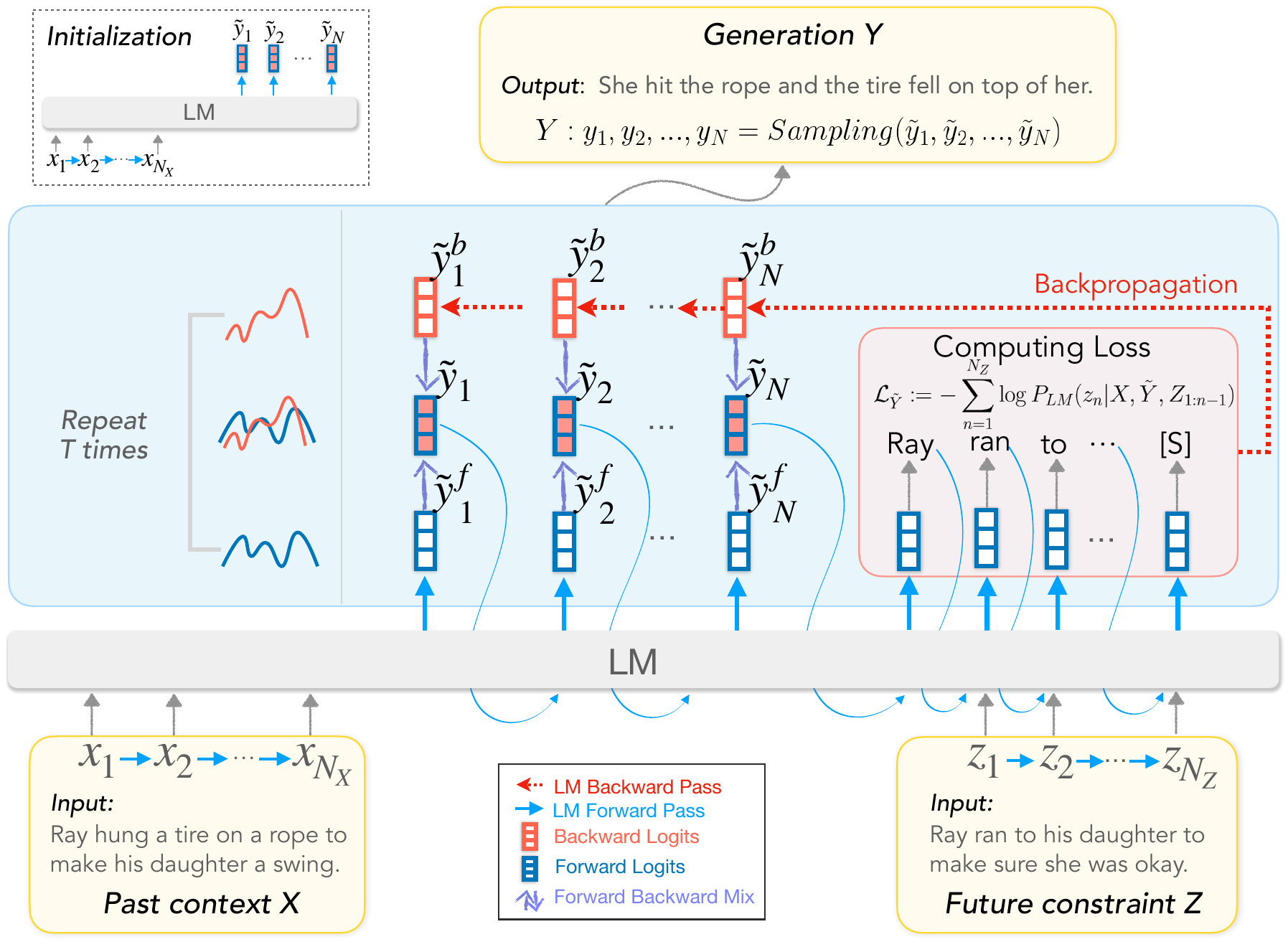}}
% \vspace{-15pt}
\caption{
%\qin{Will change}
Illustration of the \modelname decoding procedure, using abductive reasoning as an example. At initialization (upper-left box), the language model (LM) initializes the logits 
\mixvector~$\Tilde{Y}=\{\Tilde{y}_1,\dots,\Tilde{y}_N\}$ of the hypothesis by reading the past context $X$ and generating a continuation with regular decoding. At each forward-backward iteration, we compute the task-specific {\bf loss} $\mathcal{L}_{\Tilde{Y}}$ of the logits \mixvector~based on the future constraint $Z$ ({\color{red} red box}). The {\bf backward pass} then performs back-propagation and produces the backward logits \bwvector~$\Tilde{Y}^b=\{\Tilde{y}_1^b,\dots,\Tilde{y}_N^b\}$. In the subsequent {\bf forward pass}, for each step $n$, we compute the forward logits \fwvector~$\Tilde{y}_n^f$ conditioning on the preceding logits \mixvector~$\Tilde{y}_{n-1}$, and then mix it with the respective backward logits \bwvector~to produce the new logits \mixvector~$\Tilde{y}_{n}$ at step $n$.
%
%{\bf (1):} Initialize the logits $\Tilde{Y}^{(0)}$ by feeding context $X$ and generating a continuation with regular decoding. {\bf (2):} The \textit{backward} pass at iteration $t$. Compute the task-specific loss $\mathcal{L}$ using the soft representation $\Tilde{Y}^{(t-1)}$ ({\color{red} red} circle) which is then updated by backpropagation, resulting in the new backward logits $\Tilde{Y}^{(t),\text{bwd}}$ ({\color{orange} yellow} triangle). {\bf (3):} The subsequent \textit{forward} pass. For each step, compute the forward logits $\Tilde{Y}^{(t),\text{fwd}}$ ({\color{green} green} rhombus) by conditioning on $X$ and previously computed logits, and mix with the respective backward logits to produce the final logits at iteration $t$.
}
\label{fig:fig_2} 
\end{figure*}

%% file: 2-background.tex
\section{Background}
\label{sec:background}

% Most NLP benchmarks test the ability to draw conclusions from already available information. For instance, in the popular natural language inference task \cite[NLU;][]{bowman2015large}, the goal is to determine whether a hypothesis is deductively followed from a given premise. The earlier version of the task specified that the premise may be enhanced with common background knowledge \cite{dagan2013recognizing}, but no unconfirmed assumptions are allowed. When such assumptions are required, the sentence-pair should be classified as ``neutral'', i.e. there is not enough information to determine whether the hypothesis is true or not. Conversely, human reasoning involves making assumptions where information is missing. When assumptions are made, the hypothesis is inferred conditionally, and a new premise might invalidate it \cite{kraus1990nonmonotonic}. This is only feasible in nonmonotonic logic, as it contradicts the monotonicity property according to which valid arguments cannot be made invalid by adding new premises. We study two tasks of that nature: abductive reasoning (\S~\ref{ssec:background:abductive}) and counterfactual reasoning (\S~\ref{ssec:background:counterfactual}). 

Most NLP benchmarks 
%test the ability to draw conclusions from already available information. 
have focused on reasoning about information that is \emph{entailed} from the premise. 
For instance, natural language inference  \cite[NLI;][]{bowman2015large} 
focuses primarily on whether a hypothesis is 
entailed from a given premise, which means the information stated in the hypothesis is a subset of the information provided in the premise. 
%While the early task definition allowed reliance on common background knowledge \cite{dagan2013recognizing}, enhancing the premise with non-trivial assumptions classifies the sentence-pair as ``neutral''. 
However, 
%humans reason by making assumptions where information is missing. 
it has been noted that human reasoning is often the other way, where hypotheses often contain new information that was not available in the premise, but plausibly true (but possibly defeasible with new additional context) \cite{johnson2006we, mercier2017enigma}. 
%When assumptions are made, the hypothesis is inferred conditionally, and new information might invalidate it \cite{kraus1990nonmonotonic}. 
This type of reasoning corresponds to nonmonotonic reasoning \cite{kraus1990nonmonotonic}, as it contradicts the monotonicity property according to which valid arguments cannot be made invalid by adding premises. We study two tasks of that nature: abductive reasoning (\S\ref{ssec:background:abductive}) and counterfactual reasoning (\S\ref{ssec:background:counterfactual}). 

\subsection{Abductive Reasoning}
\label{ssec:background:abductive}

% Abductive reasoning aims at finding the simplest explanation to the evidence \cite{peirce1960collected}. It has a central role in human ability to ``read between the lines''. In first language acquisition, it facilitates inferring grammar rules from utterances \cite{andersen1973abductive}. In the process of understanding sentences in discourse, abductive reasoning provides the best explanation of why the sentence would be true \cite{hobbs1993interpretation}. Despite the abundance of abductive reasoning in human language understanding, relatively little focus has been given to abductive reasoning in NLP.

Abductive reasoning aims at finding the most likely explanation to partial observations  \cite{peirce1960collected}. It has a central role in the human ability to ``read between the lines,'' and is crucial for language acquisition \cite{andersen1973abductive}, understanding sentences in discourse \cite{hobbs1993interpretation}, and many more. Despite the importance, however, relatively little focus has been given to it in NLP research. 

% Recently, \newcite{bhagavatula2019abductive} proposed the abductive reasoning task. Given two observations, the goal is to determine the most likely explanation of what happened in-between. The task has two variants: abductive NLI, framed as a multiple-choice question answering task, and abductive NLG, on which we focus in this paper, which is a conditional generation task for explaining given observations in natural language.

Recently, \newcite{bhagavatula2019abductive} propose the abductive reasoning task. Given two observations, the goal is to determine the most likely explanation of what happened in-between. The dataset introduced for the task, \dataset{}, consists of 20k observations derived from the first and last sentence of stories in the ROCStories dataset \cite{mostafazadeh2016corpus}. %, with pairs of plausible and implausible explanations. %200k plausible and implausible explanations for what happened in-between were crowdsourced, and filtered to remove trivial instances and annotation artifacts \cite{Zellers2018SWAGAL}. 
We focus on the abductive NLG setup introduced in the paper, which is framed as a conditional generation task where a plausible explanation to the observations must be generated using language.
% The \dataset{} dataset was introduced for the task, consisting of over 20k observations derived from the first and last sentence of stories in the ROCStories dataset  \cite{mostafazadeh2016corpus}. 200k plausible and implausible explanations for what happened in-between were crowdsourced. These explanations were then adversarially filtered to remove trivial instances and annotation artifacts \cite{Zellers2018SWAGAL}. 
% \newcite{bhagavatula2019abductive} reported the performance of several baselines for each of the task variants. The NLI models assumed different types of inter-dependencies between the explanation and the observations, and were based on different neural building blocks (word embeddings, pre-trained LMs, etc.). The NLG models were based on pre-trained LMs, including the \comet~model, which was trained on commonsense if-then triplets from the ATOMIC knowledge base \cite{bosselut-etal-2019-comet,Sap2019ATOMICAA}. While the NLI models showed strong performance, the NLG task proved to be more challenging.
The authors reported the performance of several pre-trained LM-based baselines %, including a knowledge-informed model based on \comet{}, a LM trained on commonsense knowledge tuples from the ATOMIC knowledge base \cite{bosselut-etal-2019-comet,Sap2019ATOMICAA},
and showed promises and limitations of such approaches. %these models could generally not produce viable abductive interpretations of the observations. 

%\text{fwd}} + (1-\gamma)\cdot \Tilde{\bm{y}}_n^{(t),\text{bwd}},

\subsection{Counterfactual Reasoning}
\label{ssec:background:counterfactual}

Counterfactual reasoning aims at inferring alternative past events that could have happened given a certain change in conditions \cite{goodman1947problem,sep-counterfactuals}. While counterfactual reasoning plays an important role in AI systems \cite{isard1974would,ginsberg1986counterfactuals}, it requires causal reasoning abilities, which are arguably absent from current association-based AI \cite{pearl2018book}. While there has been work on counterfactual reasoning in NLP, including recognizing counterfactuals in text \cite{son2017recognizing}, and improving the performance of NLP tasks using \textit{counterfactual learning} %, a learning approach where the predictions of a historic system supervise the training 
\cite{lawrence2017counterfactual,lawrence2018improving}, it remains a major research challenge. % niche topic. 

% There has been rather little work on counterfactuals in NLP. \newcite{son2017recognizing} learned to recognize counterfactuals in text using a combination of rules and a bag-of-words classifier. In another line of work, researchers improved machine translation and semantic parsing using ``counterfactual learning'', a machine learning approach in which the supervision comes from the predictions of a historic system \cite{lawrence2017counterfactual,lawrence2018improving}. 

Recently, \newcite{qin2019counterfactual} introduce the task of counterfactual story generation. Given a 5-sentence original story, and an alternative context in which the second sentence of the story was altered by a counterfactual, the task is to generate a new 3-sentence story ending that addresses the alternative beginning while minimally editing the original ending. The associated \dataName~dataset is based on fictional narratives from ROCStories, for which counterfactual contexts and alternative endings are crowdsourced, % in two steps. %First, workers were shown a story and asked to write counterfactual contexts. Then, for a subset of the data, workers were asked to minimally revise the original ending such that it fit the counterfactual context. 
yielding 29,849 problem instances.
%, as well as additional 80,115 instances for which only the counterfactual context is available. %, useful for unsupervised setups.
\newcite{qin2019counterfactual} report several baseline performances, and find that models based on pre-trained LMs produce output that recognize the counterfactual, but generated endings which deviated considerably from the original storyline. In contrast, in the supervised setup, models optimize the easier of the two goals and generate endings that are overly similar to the original endings.

%% file: 3-approach.tex
% \begin{figure*}[t]
% \centering
% {\includegraphics[scale=0.23]{emnlp2020-templates/figure/figure2.pdf}}
% \vspace{-5pt}
% \caption{Illustration of the \modelname decoding procedure. {\bf (1):} Initialize the logits $\Tilde{Y}^{(0)}$ by feeding context $X$ and generating a continuation with regular decoding. {\bf (2):} Backward pass at iteration $t$. Compute the task-specific loss $\mathcal{L}$ using the soft representation $\Tilde{Y}^{(t-1)}$ ({\color{red} red } circle) which is then updated by backpropagation, resulting in the new backward logits ({\color{orange} yellow} triangle). {\bf (3):} The subsequent forward pass. For each step, compute the forward logits ({\color{green} green} rhombus) conditioning on $X$ and previously computed logits, and mix with the respective backward logits to produce the final logits at iteration $t$.}
% \label{fig:fig_2} 
% \end{figure*}

% \section{Unsupervised Reasoning \& Generation Framework}
\section{The \modelname~Approach}
\label{sec:framework}

Humans make inferences based on available information and refine them when new information arrives. Since currently available pre-trained LMs generate text by sequentially predicting the next token from left to right, they are incapable of conditioning on future constraints. Therefore, we propose \modelname: an unsupervised backprop-based decoding algorithm, which is summarized in Algorithm~\ref{alg:general}, illustrated in Figure~\ref{fig:fig_2}, and detailed below. \modelname~intermittently refines the predictions to cohere with either the context or the constraints (Section~\ref{ssec:framework:decoding_strategy}). The candidate generations are then ranked by coherence (Section~\ref{ssec:framework:ranking}). 

\subsection{Decoding Strategy}
\label{ssec:framework:decoding_strategy}
%\qin{will fix some bug soon}
% setting
Given context text $X$, the goal is to generate continuation text $Y=(y_1, \dots, y_N)$, such that $Y$ satisfies certain constraints according to the reasoning tasks, usually defined based on another context $Z$ (see Figure~\ref{fig:main_example}; we discuss the task-specific constraints in the respective task sections). 

%Note that, differing from previous open-ended text generation where the generated text is only required to be fluent, here the reasoning tasks impose additional constraints, requiring the model to actually perform reasoning in order to satisfy the constraints. \vered{This paragraph might be redundant.}

% overview

The proposed approach interleaves two procedures, namely, \emph{forward} and \emph{backward}, that produce and iteratively refine the generation, for a predefined number of iterations $T$. In particular, the \emph{forward} pass ensures the generated text is a fluent continuation of the context $X$, while the \emph{backward} pass informs the model about the constraint and steers the generation to satisfy it.  %\qin{ is this paragraph duplicated as we have mentioned something similar in the intro?}

As detailed below, the backward pass uses gradient descent to update the generation $Y$. However, $Y$ is a discrete text that is not differentiable. Instead, throughout the algorithm, we maintain a soft representation of the sequence $\Tilde{Y}=(\Tilde{\bm{y}}_1, \dots, \Tilde{\bm{y}}_N)$, where $\Tilde{\bm{y}}_n\in\mathbb{R}^{V}$ represents the logits of the $n$-th token and $V$ is the vocabulary size. After the logits are refined over multiple iterations of the forward and backward passes, we generate discrete text at each step by sampling from $y_n \sim \text{softmax}(\Tilde{\bm{y}}_n / \tau)$, where $\tau>0$ is the temperature.

\input{emnlp2020-templates/figure/algorithm}

% We start by initializing the logits $\Tilde{Y}$. As shown in Figure~\ref{fig:fig_2}~(1), we feed the context $X$ into the LM, and generate a continuation of length $N$ with regular decoding (e.g., greedy). We set the output logits of the continuation as the initialization, denoted as $\Tilde{Y}^{(0)}$.

We start by initializing the logits before the first iteration, $\Tilde{Y}^{(0)} = (\Tilde{y}_1^{(0)} \dots \Tilde{y}_N^{(0)})$, by feeding the context $X$ into the LM and greedily decoding $N$ continuation tokens. 
%(left box in Figure~\ref{fig:fig_2}).

% 
% \paragraph{Backward}
% In order to steer the generation to satisfy the constraint, the backward pass uses gradient backpropagation to update the generation. Specifically, given a reasoning task, we could express the constraint as a loss function $\mathcal{L}(X, {\Tilde{Y}}^{(t-1)}, Z)$ which evaluates how well the generation $Y$ (approximated with the soft representation $\Tilde{Y}$) obeys the constraint. (Concrete instantiations of the loss are described in the subsequent sections.)
% The goal of this pass is to minimize the loss w.r.t the generation, as shown in Figure~\ref{fig:fig_2}~(2).
% Specifically, at iteration $t$, for each step $n$ in the generation, we update its logits with:
\paragraph{Backward}
The backward pass uses gradient backpropagation to update the generation with respect to the constraint. Specifically, we express the task-specific constraint as a loss function $\mathcal{L}(X, {\Tilde{Y}}^{(t-1)}, Z)$ that evaluates how well the generation $Y$ (approximated with the soft representation $\Tilde{Y}$) obeys the constraint (see the subsequent sections for concrete instantiations of the loss). The goal of this pass is thus to minimize the loss w.r.t the generation. Specifically, at iteration $t$, for each step $n$ in the generation, we update its logits with:
\begin{equation}
\resizebox{0.89\hsize}{!}{%
% \begin{split}
$\Tilde{\bm{y}}_n^{(t),\text{b}} = \Tilde{\bm{y}}_n^{(t-1)} - \lambda \cdot \nabla_{\Tilde{\bm{y}}_n} \mathcal{L}(X, {\Tilde{Y}}^{(t-1)}, Z),$ %
}
% \end{split}
\label{eq:bwd-logits}
\end{equation}
where $\nabla_{\Tilde{\bm{y}}_n} \mathcal{L}(X,\Tilde{Y}^{(t-1)},Z)$ is the gradient of the constraint-informed loss $ \mathcal{L}$ w.r.t the $n$-th logits, and $\lambda\in\mathbb{R}$ is the step size. In practice, we may repeat the gradient updates multiple times in a single pass. 

%Given a pre-defined length $N$, all tokens from step 1 to $N$ will have their logits updated in the backward pass. The forward pass introduced in the following could go beyond length $N$ to continue generation and ensure producing complete text.

% \paragraph{Forward} 
% We then perform a forward pass that refines the generation to ensure the text is fluent and coherent with the preceding context $X$. The pass at each step $n$ consists of one step left-to-right computation and a mixing operation. Specifically, at iteration $t$, for a particular step $n$, we first compute the forward logits with the LM:
\paragraph{Forward} The forward pass ensures that $Y$ is fluent and coherent with the preceding context $X$. At iteration $t$, for a particular step $n$, we compute the forward logits with the LM:
\vspace{-7pt}
\begin{equation}
%\small
% \begin{split}
% \Tilde{\bm{y}}_n^{(t),\text{fwd}} = \text{LM}(X, \Tilde{Y}_{1:n-1}^{(t)} / \tau).
\Tilde{\bm{y}}_n^{(t),\text{f}} = \text{LM}(X, \Tilde{Y}_{1:n-1}^{(t)}).
% \end{split}
\label{eq:fwd-logits}
\end{equation}
We then \emph{mix} the $n$th-step forward and backward logits to get the final logits of iteration $t$:
\begin{equation}
%\small
\begin{split}
\Tilde{\bm{y}}_n^{(t)} = \gamma\cdot \Tilde{\bm{y}}_n^{(t),\text{f}} + (1-\gamma)\cdot \Tilde{\bm{y}}_n^{(t),\text{b}},
\end{split}
\label{eq:combined-logits}
\end{equation}
where $0<\gamma<1$ is the mixing weight. The resulting logits $\Tilde{\bm{y}}_n^{(t)}$ are then fed to the LM to compute the forward logits at the $(n+1)$th step (Eq.\ref{eq:fwd-logits}). This way, information from the backward pass is integrated into the left-to-right generation process to produce text that is informed by the constraint.

% \begin{equation}
% \begin{split}
% \nabla_{\Tilde{Y}} \mathcal{L}(\Tilde{Y})
% \end{split}
% \end{equation}

% \paragraph{Forward}
% The forward pass refines the generation to ensure the text is fluent and coherent with the preceding context $X$. We compute the forward logits with a left-to-right pass conditioning on the context, and mix them with the current backward logits. More specifically, at iteration $t$, for each step $n$, we first compute the forward logits with the LM:
% \begin{equation}
% \begin{split}
% \Tilde{\bm{y}}_n^{(t),\text{fwd}} = \text{LM}(\Tilde{Y}_{1:n-1}^{(t)} / \tau).
% \end{split}
% \label{eq:fwd-logits}
% \end{equation}
% We then combine the forward logits with the backward logits we obtained in Eq.\eqref{eq:bwd-logits} to get the resulting logits of iteration $t$:
% \begin{equation}
% \begin{split}
% \Tilde{\bm{y}}_n^{(t)} = \gamma\cdot \Tilde{\bm{y}}_n^{(t),\text{fwd}} + (1-\gamma)\cdot \Tilde{\bm{y}}_n^{(t),\text{bwd}},
% \end{split}
% \label{eq:combined-logits}
% \end{equation}
% where $0<\gamma<1$ is the mixing weight. The resulting logits are $\Tilde{\bm{y}}_n^{(t)}$ then fed into the LM to compute the forward logits at the next step. This way, information from the backward pass is integrated into the left-to-right generation process to produce text that is informed by the constraint.

We pre-define the number of tokens $N$ required by the backward pass, but we allow the forward pass to generate more than $N$ tokens if those are needed to obtain complete sentences. In that case, we set the logits of the extra tokens to the forward logits, without mixing: $
\Tilde{\bm{y}}_n^{(t)} = \Tilde{\bm{y}}_n^{(t),\text{f}}  ~~\text{for}~~n>N$. We then prune any trailing tokens in the sampled text to get complete sentences.

\subsection{Ranking}
\label{ssec:framework:ranking}

% The output of the decoding step is a list of candidate generations for each iterations: $Y_s = \{Y^{(t)} | t = 1, ..., T\}$. We further use an unsupervised approach to rank and pick the best sample as the final output. Specifically, we take advantage of the BERT model, which was pre-trained with a next-sentence prediction (NSP) objective. Given two sentences $A$ and $B$, we use BERT to compute the coherence between the two sentences, i.e., the likelihood of $B$ following $A$:
The output of the decoding step is a list of candidate generations for each iteration: $Y_s = \{Y^{(t)} | t = 1, ..., T\}$. We further use an unsupervised approach to rank and pick the best sample as the final output. Specifically, we take advantage of the BERT model, which was pre-trained with a next-sentence prediction (NSP) objective. Given two sentences $A$ and $B$, we use NSP to compute the likelihood of $B$ following $A$ as a proxy for coherence:
% The generation and refinement approach in Section~\ref{ssec:framework:decoding_strategy} generates different samples at each iteration. We further use an unsupervised approach to rank and pick the best sample as the final output. Specifically, we take advantage of the BERT model, which was pre-trained with an next-sentence prediction (NSP) objective. Given two sentences $A$ and $B$, we use BERT to compute the coherence between the two sentences, i.e., the likelihood of $B$ following $A$:
\begin{equation}
\begin{split}
c(A, B) = \text{BERT\_NSP}(A, B),
\end{split}
\end{equation}
where $c(\cdot,\cdot)$ denotes the coherence score. %As we will see in the following sections, 
This score is used to evaluate the quality of a given candidate continuation $Y$ by measuring (1) its compatibility with the subsequent text of the context $X$, (2) the internal consistency of $Y$ if it consists of multiple sentences, and (3) the compatibility of $Y$ with its right-side text when it is applicable.

% The above generation and refinement approach could generate multiple samples throughout the iterations. We further use an unsupervised approach to rank and pick the best sample as the final output. Specifically, we take advantage of the BERT model, which was pre-trained with an next-sentence prediction (NSP) objective. Given two sentences $A$ and $B$, we use BERT to compute the consistency between the two sentences, i.e., the likelihood of $B$ following $A$:
% \begin{equation}
% \begin{split}
% C(A, B) = \text{BERT\_NSP}(A, B),
% \end{split}
% \end{equation}
% where $C(\cdot,\cdot)$ denotes the consistency score.
% With this capability, for each candidate, we can evaluate the compatibility of the candidate $Y$ as subsequent text of the context $X$, internal consistency of $Y$ if it consists of multiple sentences, as well as the compatibility of $Y$ with its right-side text if any.

%% file: emnlp2020-templates/figure/algorithm.tex
% % Algorithm
\begin{algorithm}[!t]
\small
\centering
\caption{\small \modelname Decoding}
\label{alg:general}
\begin{algorithmic}[1]
\REQUIRE Pre-trained language model (LM) \\
\quad\ \  Context $X$ \\
\quad\ \  Future constraint $Z$\\
%\quad\ \  Context $X$ and other info (e.g., $Z$) in the instance\\
% \quad\ \ Constraint expressed as a loss function $\mathcal{L}(\cdot)$ \\ 
\STATE Initialize logits $\Tilde{Y}^{(0)}$
\STATE Initialize $Y_s$, list of candidate generations
\FOR {$t \gets 1$ to $T$}
    \STATE {\color{gray}// Backward pass}
    \FOR {$n \gets N$ to $1$} 
        \STATE Compute backward logits $\Tilde{\bm{y}}^{\text{b}}_n$, Eq.\eqref{eq:bwd-logits}
    \ENDFOR
    \STATE {\color{gray}// Forward pass}
    \FOR {$n \gets 1$ to $N$} 
        \STATE Compute forward logits $\Tilde{\bm{y}}^{\text{f}}_n$, Eq.\eqref{eq:fwd-logits}
        \STATE Mix forward and backward logits, Eq.\eqref{eq:combined-logits}
    \ENDFOR
    \STATE Sample candidate $Y$ from logits $\Tilde{Y}$ and add to $Y_s$
\ENDFOR  
\STATE Rank $Y_s$ by coherence
\ENSURE The most coherent generated text $Y$ from $Y_s$
\end{algorithmic}
\end{algorithm}

% % Algorithm
% \begin{algorithm}[!t]
% \small
% \centering
% \caption{\scriptsize Unsupervised Reasoning \& Generation}
% \label{alg:general}
% \begin{algorithmic}[1]
% \REQUIRE Pre-trained language model (LM) \\
% \quad\ \  Context $X$ \\
% \quad\ \  Additional information $Z$\\
% % \quad\ \ Constraint expressed as a loss function $\mathcal{L}(\cdot)$ \\ 
% \STATE Initialize $Y_s$, list of candidate generations
% \FOR {$t \gets 1$ to $T$}
%     \FOR {$n \gets 1$ to $N$} 
%         \STATE $\Tilde{\bm{y}}^{(t),\text{fwd}}_n \leftarrow \operatorname{forward\_pass}(LM, X)$ {\color{gray} // Eq.\eqref{eq:fwd-logits}}
%     \ENDFOR
    
%     \FOR {$n \gets N$ to $1$} 
%         \STATE $\Tilde{\bm{y}}^{(t),\text{bwd}}_n \leftarrow \operatorname{backward\_pass}(LM, Z)$  {\color{gray} // Eq.\eqref{eq:bwd-logits}}
        
%         \STATE $\Tilde{\bm{y}}^{(t)}_n \leftarrow \operatorname{mix}(\Tilde{\bm{y}}^{(t),\text{fwd}}_n, \Tilde{\bm{y}}^{(t),\text{bwd}}_n)$ {\color{gray} // Eq.\eqref{eq:combined-logits}}
%     \ENDFOR
    
%     \STATE Sample candidate $Y^{(t)}$ from logits $\Tilde{Y}^{(t)}$ 
%     \STATE $Y_s.\operatorname{add}(Y^{(t)})$
% \ENDFOR  
% \STATE Rank $Y_s$ by coherence
% \ENSURE The most coherent generated text $Y$ from $Y_s$
% \end{algorithmic}
% \end{algorithm}

%% file: 4-abductive.tex
\section{Task 1: Abductive Reasoning}
\label{sec:tasks:abductive}

\begin{figure*}[t]
\centering
{\includegraphics[scale=0.4]{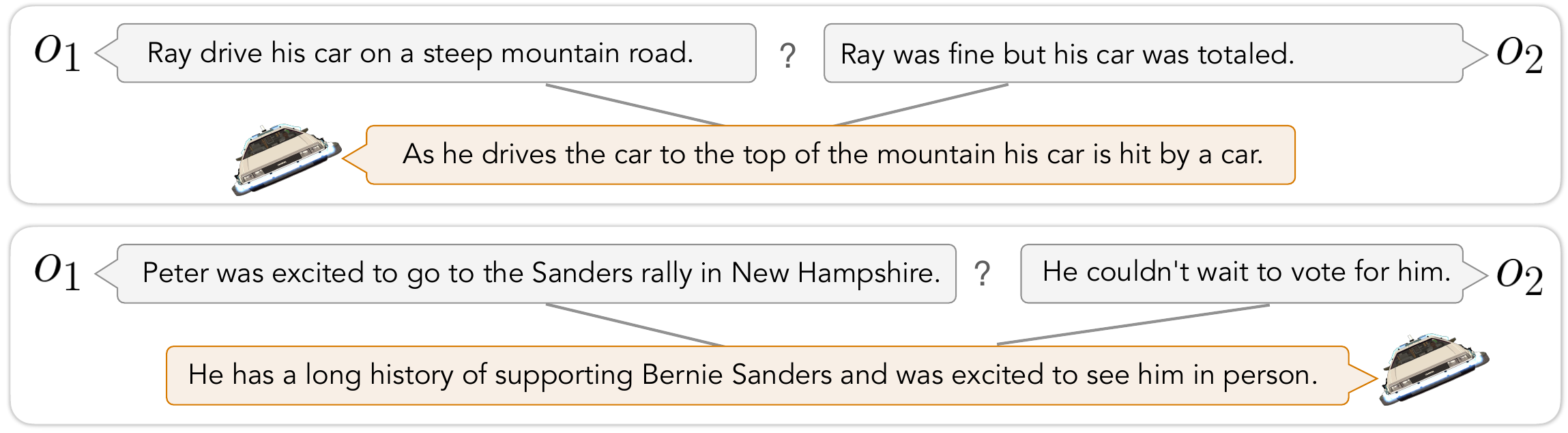}}
% \vspace{-15pt}
\caption{
Examples of generated hypotheses on three abductive reasoning cases. 
Given observations O1 and O2, \modelname generates a hypothesis explaining the observations.}
\label{fig:fig_exp_abd} 
\end{figure*}

% We are given an Observation\_1 $O_1$ as well as an Observation\_2 $Z$. The task is to generate an explanation $O_2$ such that $(X,Y,Z)$ forms a coherent story. In other words, $Y$ must take into account both the left-side context $X$ and the right-side context $Z$.
Each instance in the \dataset{} dataset consists of two observations $O_1$, $O_2$ and a hypothesis $H$ that explains the two observations. These inputs naturally map to $X$, $Z$ and $Y$ in our framework. Formally, the abductive generation task aims to maximize {$P(Y|X,Z)$} -- i.e. models must consider both left and right contexts ($X$ and $Z$) jointly.
% The task is to generate a plausible explanation Y given two observations X, Z such that $(X,Y,Z)$ forms a coherent story. In other words, $Y$ must take into account both the left-side context $X$ and the right-side context $Z$. Formally, the task aims to maximize {$P(Y|X,Z)$}.

\input{emnlp2020-templates/figure/abductive_automatic}

\subsection{Task Setup}
\label{sec:tasks:abductive:task_setup}

\paragraph{Constraints} 
% While we would like to maximize $Y$ given $Z$, it is not feasible to maximize $Y$ given a prefix $Z_{1:n}$ of $Z$ which is likely not a complete sentence. We thus reverse the order and maximize $Z$ given $Y$, assuming that this objective still accounts for the coherence between the two sentences. We define the loss function as the cross-entropy loss of generating $Z$ given $Y$ with the LM:

We maximize $Z$ given $X\Tilde{Y}$ by defining the loss function as the cross-entropy loss of generating $Z$ given $X\Tilde{Y}$ with the LM:\footnote{Note that this is applied to each prefix of $\Tilde{Y}$, although some of them are not complete sentences.}
\begin{equation}
\resizebox{0.89\hsize}{!}{%
$\mathcal{L}(X,\Tilde{Y},Z) := -\sum_{n=1}^{N_{Z}} \log P_{\LM}(z_n | X,\Tilde{Y}, Z_{1:n-1}),$
}
\label{eq:abd-loss}
\end{equation}
where $P_{\LM}(a_j | a_{1:j-1})$ is the likelihood of generating token $a_j$ given the preceding text $a_{1:j-1}$.

Following the earlier study of the task~\citep{bhagavatula2019abductive}, we also prepend $Z$ to $X$ to ``leak'' the future information to the LM. That is, we replace $X$ with $Z\langle e \rangle X$ in the above equation, where $\langle e\rangle$ denotes a special end-of-text token. However, the comparisons with respective baselines below show the prepended $Z$ is minor to the performance.

% \begin{equation}
% \begin{split}
% \mathcal{L}(X,\Tilde{Y},Z) :=& - \log P_{\LM}(Z|X,\Tilde{Y}) \\
% =& - \sum_{n=1}^{N_{Z}} \log P_{\LM}(z_n | X,\Tilde{Y}, Z_{1:n-1}),
% \end{split}
% \label{eq:abd-loss}
% \end{equation}
% where $P_{\LM}(z_n | X, \Tilde{Y}, Z_{1:n-1})$ is the likelihood of generating token $z_n$ by the LM given the preceding text $(X, \Tilde{Y}, Z_{1:n-1})$.

\paragraph{Ranking}
We rank candidates by the overall coherence after inserting $Y$ in between $X$ and $Z$:
\begin{equation}
% \begin{split}
\resizebox{0.89\hsize}{!}{%
$\operatorname{ranking\_score}(Y) = c(XY, Z) + c(X, YZ).$%
}
% \end{split}
\label{eq:abd-ranking}
\end{equation}
% That is, we concatenate $X$ and $Y$ as a single piece of text and evaluate its coherence with the subsequent $Z$, and similarly, we concatenate $Y$ and $Z$ and evaluate the coherence between $X$ and $YZ$.

% \begin{equation}
% \begin{split}
% \mathcal{L}(\Tilde{Y}) :=& - 2 \log P_{\LM}(Z|\Tilde{Y}, X) \\
% & + \log P_{\LM}(Z|\Tilde{Y}) \\
% & + \log P_{\LM}(Z|X)
% \end{split}
% \end{equation}

% \paragraph{Hyperparameters} 
% %While these models originally use the small \textbf{GPT2-117M} \cite{radford2019language} as a pretrained starting point, we replace their initialization with \textbf{GPT2-345M} to be in line with our own approach. 
% For all models, we use \textbf{GPT2-345M} as the pre-trained LM for reasoning and generation. We use the \dataset{} development set to select hyperparameters. For our approach, we set the hypothesis length $N=15$ in the backward pass and allow the forward pass to generate 30 tokens for complete sentences \antoine{This seems to imply that $N$ is more for the forward pass than backward pass? Maybe we should define this with two separate variables?}. We run $T=20$ forward-backward iterations, with each backward pass performing $20$ gradient updates using a small step size $\lambda=0.0003$ \antoine{we perform 20 gradient updates during each backward pass? There's more than one update per iteration?}. The mixing weight of forward/backward logits is $\gamma=0.88$. We use greedy decoding to produce a single candidate at each iteration $T$. We include the code in the supplementary materials which will be cleaned and released upon acceptance.

\paragraph{Hyperparameters} We use GPT2-345M \cite{radford2019language} as the pre-trained LM for all models. We use the \dataset{} development set to select hyperparameters. %We run $T=20$ forward-backward iterations and use greedy decoding for our method. We found that the baselines generally perform best with top k decoding \cite{fan2018hierarchical} ($k=40, \tau=0.7$). Other hyperparameters are outlined in Appendix~\ref{app:task:counterfactual}. %We include the code in the supplementary materials which will be cleaned and released upon acceptance. \antoine{Maybe we should mention code release in the intro?}
We use greedy decoding for our method and top k decoding \cite{fan2018hierarchical} ($k=40, \tau=0.7$) for our baselines. Other hyperparameters are outlined in Appendix~\ref{app:task:abductive}.

\subsection{Experimental Setup}
\label{sec:tasks:abductive:experimental_setup}

\paragraph{Baselines} 
We compare our method against baselines from \newcite{bhagavatula2019abductive}. The unsupervised baselines use a pre-trained GPT-2 model to generate $Y$ given a prompt text---either the observation $X$ alone (\zs$_X$) or $Z\langle e \rangle X$ (\zs$_{ZX}$). The supervised method (Sup) follows the same input format as \zs$_{ZX}$, but fine-tunes GPT-2 on the \dataset{} training set. Finally, our knowledge-informed baseline (\lmallcometembprefix) further augments the representation of Sup with knowledge from \comet~\cite{bosselut-etal-2019-comet}. 

To separately study the contribution of our decoding strategy and ranking component, we also report the performance of ranking the baseline outputs. 
Specifically, we let each baseline generate 20 candidates and rank them by coherence (Eq.~\ref{eq:abd-ranking}).\footnote{We tried ablating the ranking component from our method in preliminary experiments, and found that ranking is essential to obtaining good performance. By adding ranking to our baselines, we assess %\qin{is this a typo?}
the contribution of our decoding strategy.}

\subsection{Results}
\label{ssec:abductive_results}

\paragraph{Automatic Evaluation}

We report the same metrics as \newcite{bhagavatula2019abductive}: BLEU-4 \cite{papineni2002bleu}, ROUGE-L \cite{lin2004rouge} and \textsc{BertScore} \cite{Zhang2019BERTScoreET} (with the {\textit{bert-base-uncased} model}). The results in Table~\ref{tab:abd_autoeval} show that \modelname performs best among the unsupervised systems across all metrics. We also note that our ranking step improves both the performance of our model and that of the zero-shot baselines. 

%\paragraph{Ablations}

% \paragraph{Metrics}
% We follow \newcite{bhagavatula2019abductive} and use automatic metrics that have shown to correlate closely with human judgments, including BLEU-4 \cite{papineni2002bleu}, ROUGE-L \cite{lin2004rouge} and \textsc{BertScore} \cite{Zhang2019BERTScoreET} (with the {\textit{bert-base-uncased} model}). 

% We also conduct two sets of human evaluations on 100 test examples through crowdsourcing on Amazon Mechanical Turk. In the calibration setting, for each example, workers were presented a pair of observations ($X$ and $Z$) and a generated hypothesis $Y$, and were asked whether the hypothesis is coherent with the observation $X$ ($X$-$Y$), the observation $Z$ ($Y$-$Z$), and both ($X$-$Y$-$Z$), using a 4-point Likert scale. In the pairwise comparison setting, we compared our approach with each of the other baselines. Workers were presented the outputs of a pair of systems and asked to choose which output is better in terms of the 3 coherence criteria detailed above. Each example was labeled by 3 workers. \qin{kappa score}

%and asked which model does better on explaining both X and Z. We evaluate 100 examples from each models and each example was labeled by three different workers. 
% ``equally good/bad''
\input{emnlp2020-templates/figure/abductive_human_calibration}
\input{emnlp2020-templates/figure/abductive_human_pairwise}

\paragraph{Human Evaluation}

We conduct two sets of human evaluations on 100 test examples using crowdworkers from Amazon Mechanical Turk. In the scoring setting, presented in Table~\ref{tab:abd_calibration}, workers were presented a pair of observations ($X$ and $Z$) and a generated hypothesis $Y$, and asked to rate the coherence of the hypothesis with respect to the observation $X$ ($X$-$Y$), the observation $Z$ ($Y$-$Z$), and both ($X$-$Y$-$Z$), on a 4-point Likert scale. In the pairwise comparison setting, presented in Table~\ref{tab:abd_compare}, workers were presented the outputs from a pair of systems (\modelname and baseline) and asked to choose the better output in terms of the same coherence criteria. Each example was labeled by 3 workers.\footnote{The average inter-rater agreement measured by Fleiss’ $\kappa = 0.44$ (“moderate agreement”) \cite{fleiss1971measuring}.}

%Tables~\ref{tab:abd_calibration} and~\ref{tab:abd_compare}  show the calibration and pairwise comparison results, respectively. 
In both evaluation setups, our method substantially outperform the unsupervised baselines, achieving a relative improvement of $36\%-215\%$ with respect to $Y$-$Z$ coherence. Our method also outperform the supervised methods with respect to $X$-$Y$ coherence (Table~\ref{tab:abd_calibration}), and achieve competitive performance in the pairwise comparison (Table~\ref{tab:abd_compare}). Again, the ranking component contributes to increasing performance for the zero-shot baselines. Finally, the large performance gap between the methods and human-written explanations stresses the difficulty of this reasoning task and warrants future research.

% . We see that, among the unsupervised methods, our approach performs significantly better than others across all the three criteria. In particular, the $X$-$Z$ coherence shows $36\%-215\%$ relative improvement over other unsupervised methods, demonstrating the effectiveness of the proposed backpropagation-based decoding for non-monotonic reasoning. The ranked \texttt{Zero-Shot$_{o1o2}$} achieves stronger results than its vanilla version, which shows our unsupervised ranking component is helpful in selecting the best candidates. It is interesting to note that our \modelname~method is not far away from the supervised methods and even outperforms on $X$-$Y$ coherence. The large performance gap between all the methods with the human-written explanations shows the difficulty of the reasoning and calls for more study in the future.

% Table~\ref{tab:abd_compare} shows the results of human pairwise comparison, where we have consistent observations. Comparing with the enhanced ranked unsupervised baselines, our approach is preferred by human annotators on more cases by considering the overall quality.

\paragraph{Qualitative Analysis} 
Figure~\ref{fig:fig_exp_abd} presents two example outputs produced by \modelname. We can see our approach generates reasonable hypotheses by taking into account both the past and future contexts. For instance, in the first example, the future observation (O2) ``car was totaled'' indicates that Ray had a car accident, which is correctly captured in the generated hypothesis ``car is hit by a car''.
% Finally, for qualitative analysis, Table~\ref{tab:abd_sample} gives samples from our approach produced at different backward-forward iterations, along with the ranking score evaluated with Eq.\eqref{eq:abd-ranking}. We can see that the generation at the 1st iteration seems totally irrelevant to the observation O2, while that of the 2nd iteration is more relevant in terms of sentiment (e.g., word ``However'') though semantically still not coherent. As the reasoning iterations go on, more coherent samples are generated at the subsequent iterations, with high ranking scores. The ranker correctly picks the 5th candidate which is a high-quality hypothesis.

% \input{emnlp2020-templates/figure/abductive_human_calibration}
% \input{emnlp2020-templates/figure/abductive_human_pairwise}
% \input{emnlp2020-templates/figure/abductive_automatic}
% \input{emnlp2020-templates/figure/abductive_sampling_table}

%% file: emnlp2020-templates/figure/abductive_automatic.tex
\begin{table}[t]
\small
    \centering
    \begin{tabular}{lccc}
    \cmidrule[\heavyrulewidth]{1-4}
      Model & BLEU-4 & ROUGE-L & BERT \\ 
    \cmidrule[\heavyrulewidth]{1-4 }  
    \rowcolor[gray]{0.95} \multicolumn{4}{l}{\it Supervised}   \\ 
    Sup     & 3.46      & 25.60      & 49.38 \\
    \lmallcometembprefix    & 4.06     & 26.06    &49.71  \\
    %  \cmidrule{1-6}
     \rowcolor[gray]{0.95} \multicolumn{4}{l}{\it Unsupervised}   \\ 
    % \cmidrule{1-6}
    \zs$_{X}$      &0.65      & 14.99      & 39.36  \\
    \zs$_{ZX}$     &0.53     & 14.23      & 40.03  \\
    \zs$_{X}$-Ranked        &0.87      & 16.76      & 41.58  \\
    \zs$_{ZX}$-Ranked     &0.98     & 17.25      & 41.93  \\
    %\cmidrule{1-4}
    \textbf{\ourmodel{}}      & \textbf{1.38}      & \textbf{18.94}      & \textbf{42.86}  \\
    \cmidrule{1-4}
    % \multicolumn{4}{c}{\it Show to human} & \multicolumn{3}{c}{$\bm{s}_1, \bm{s}_2, \bm{s}'_2, \bm{y}$} \\
    \textit{Human}                   & 8.25      & 30.40      & 53.30 \\
    \cmidrule[\heavyrulewidth]{1-4}
    \end{tabular}
    \vspace{-10pt}
    \caption{Automatic evaluation results on the abductive task, using the test set of \dataset{}.} % We highlight the best results of the unsupervised methods.} 
    \label{tab:abd_autoeval}
\end{table}

%% file: emnlp2020-templates/figure/abductive_human_calibration.tex
\begin{table}[t]
\small
    \centering
    \begin{tabular}{lrrr}
    \cmidrule[\heavyrulewidth]{1-4}
      Model & $X$-$Y$ & $Y$-$Z$ & $X$-$Y$-$Z$ \\ 
    \cmidrule[\heavyrulewidth]{1-4 }
      \rowcolor[gray]{0.95} \multicolumn{4}{l}{\it Supervised }  \\
    Sup     & 0.510     & 0.375      & 0.314 \\
    \lmallcometembprefix    & 0.466     & 0.342    &0.286  \\
    %  \cmidrule{1-6}
     \rowcolor[gray]{0.95} \multicolumn{4}{l}{\it Unsupervised}  \\ 
    % \cmidrule{1-6}
    % \zs$_{X}$      &0.233      & 0.103      & 0.108  \\
    \zs$_{ZX}$     &0.233      & 0.103      & 0.108  \\
    \zs$_{X}$-Ranked        &0.478      & 0.208      & 0.195  \\
    \zs$_{ZX}$-Ranked     &0.474     & 0.238      & 0.236  \\
     %\cmidrule{1-4}
   \textbf{\ourmodel}        & \textbf{0.522}      & \textbf{0.325}      & \textbf{0.297}  \\
    \cmidrule{1-4}
    % \multicolumn{4}{c}{\it Show to human} & \multicolumn{3}{c}{$\bm{s}_1, \bm{s}_2, \bm{s}'_2, \bm{y}$} \\
    \textit{Human}                   & 0.879      & 0.823      & 0.783 \\
    \cmidrule[\heavyrulewidth]{1-4}
    \end{tabular}
    \vspace{-8pt}
    \caption{Human calibration results on test set of \dataset{}~. All scores are normalized to $[0,1]$.} %    We highlight the best results of the unsupervised methods (including ours and the zero-shot baselines).} 
    \label{tab:abd_calibration}
\end{table}

%% file: emnlp2020-templates/figure/abductive_human_pairwise.tex
\begin{table}[t]
\small
\centering
\scalebox{0.94}{
\begin{tabular}{@{}ll c ll@{}}
\toprule
\multicolumn{5}{c}{Overall - Human Judges Preferred}                              \\ \midrule
\rowcolor[gray]{0.95} \multicolumn{2}{c}{Our model} & \multicolumn{1}{c}{Neutral} & \multicolumn{2}{c}{Comparator} \\
\ourmodel & 21\% & 43\%  & \textbf{36\%} &  Sup         \\
\ourmodel & 25\% & 44\%  & \textbf{31\%} & \lmallcometembprefix   \\
\cmidrule{1-5}
\ourmodel & \textbf{23\%} & 62\% & 15\% & \zs$_{X}$-Ranked     \\
\ourmodel& \textbf{27\%} & 50\% & 23\%  & \zs$_{XZ}$-Ranked  \\ 
\cmidrule{1-5}
\ourmodel & 3\% & 11\% & \textbf{86\%} & Human \\
 \bottomrule
\end{tabular}
}
\vspace{-5pt}
\caption{
Human pairwise comparison results on the test set of \dataset{}, between \modelname{} and each of the baselines, by jointly considering all 3 criteria from Table~\ref{tab:abd_calibration}. ``Neutral'' means ``equally good/bad''.
%Percentage of ``equally bad'' is omitted.
}
\label{tab:abd_compare}
\end{table}

% \begin{table}[t]
% \small
% \centering
% \begin{tabular}{@{}ll|c|ll@{}}
% \toprule
% \multicolumn{5}{c}{Overall - Human Judges Preferred}                              \\ \midrule
% \rowcolor[gray]{0.95} \multicolumn{2}{c|}{Our model} & \multicolumn{1}{c|}{Neutral} & \multicolumn{2}{c}{Comparator} \\
% \ourmodel & 0.21 & -  & \textbf{0.36} &  Sup         \\
% \ourmodel & 0.25 & -  & \textbf{0.31} & \lmallcometembprefix   \\
% \ourmodel & \textbf{0.23} & - & 0.15 & \zs$_{o1}$-Ranked     \\
% \ourmodel& \textbf{0.27} & - & 0.23  & \zs$_{o1o2}$-Ranked  \\ 
% \midrule
% \ourmodel & 0.03 & - & 0.86 & Human            \\
%  \bottomrule
% \vspace{0.2mm}
% \end{tabular}
% \\
% \caption{ Human pairwise comparison results on the \textbf{Abductive task}, between our best model (Best) and comparison models on all three questions. ``Neutral'' means both are ``equally good''. }
% %Percentage of ``equally bad'' is omitted.
% \label{tab:results}
% \end{table}

%% file: 5-counterfactual.tex
\section{Task 2: Counterfactual Reasoning}
\label{sec:tasks:counterfactual}

\input{emnlp2020-templates/figure/counterfactual_automatic}

\input{emnlp2020-templates/figure/counterfactual_human_calibration}
\input{emnlp2020-templates/figure/counterfactual_human_pairwise}

\vspace*{2mm}
Given an original story ending $Z$ of story context $X^{ori}$, and a counterfactual condition $X$ that changes $X^{ori}$ to invalidate $Z$ (see Fig.~\ref{fig:main_example}), the task is to generate a new story ending $Y$ that minimally edits the original ending $Z$ to regain coherence with the counterfactual condition $X$~\cite{qin2019counterfactual}.

% We are given a story ending $Z$ of an original beginning $X^{ori}$. We are further given a counterfactual context $X$ that changes the condition in $X^{ori}$ and invalidates $Z$ (see Figure~\ref{fig:main_example} for an example). The task is to generate a new story ending $Y$ that minimally edits the original ending $Z$ to regain coherence with the counterfactual condition $X$~\cite{qin2019counterfactual}.

% \input{emnlp2020-templates/figure/counterfactual_sampling_table}
\begin{figure*}[t]
\centering
{\includegraphics[scale=0.38]{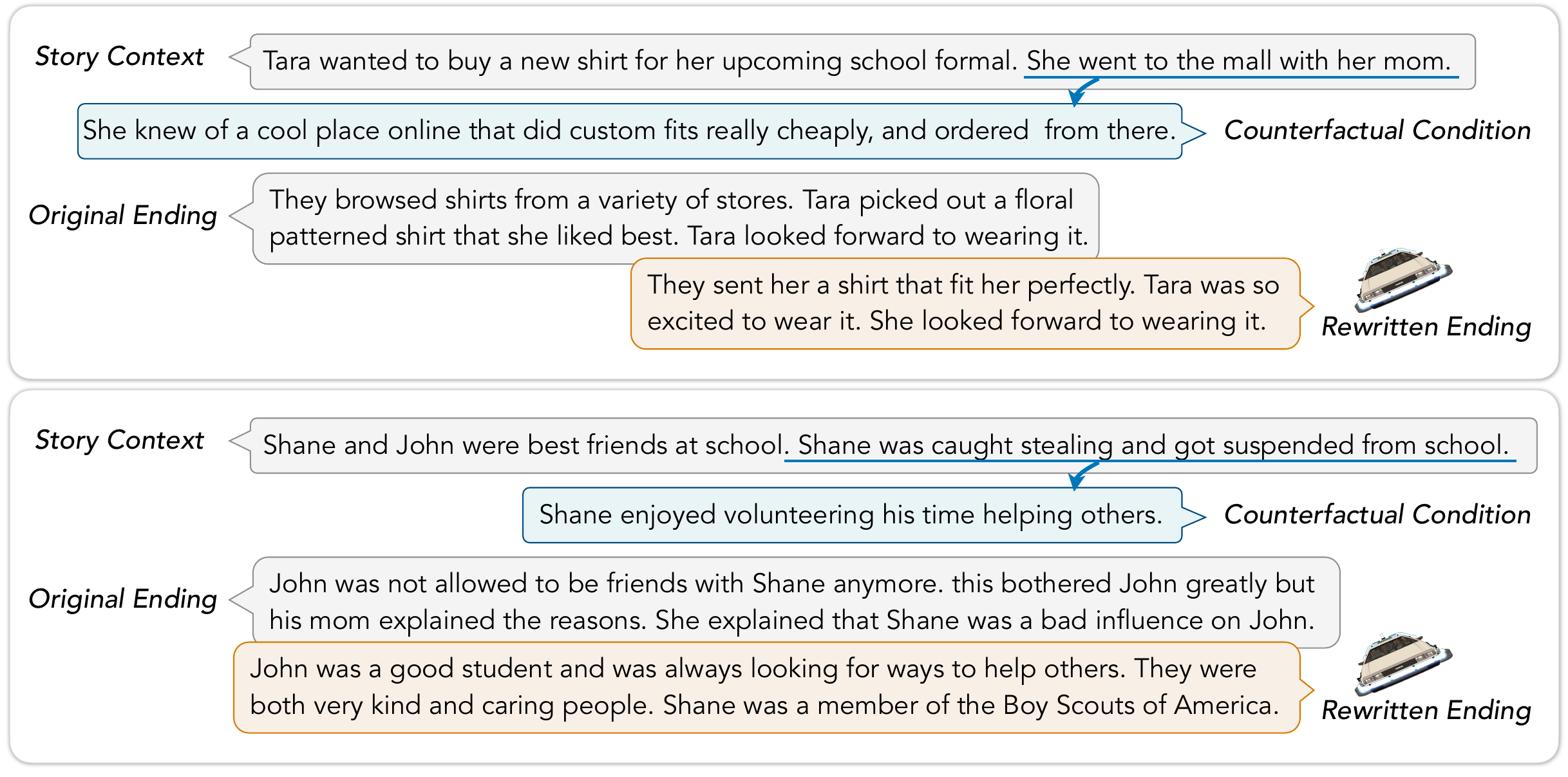}}
\vspace{-10pt}
\caption{
Examples of generated story endings on three counterfactual reasoning cases. Given a story context, a counterfactual condition, and a original ending, \modelname generates a rewritten ending which is coherent with the counterfactual condition and is similar to the original ending.}
\label{fig:fig_exp_ctf} 
\end{figure*}

\subsection{Task Setup}
\label{sec:tasks:counterfactual:experimental_setup}

\paragraph{Constraints}
The constraint we enforce is that $Y$ is close to $Z$ (i.e., minimal edits). We impose this constraint by minimizing their KL divergence:
\begin{equation}
\begin{split}
\mathcal{L}(X,\Tilde{Y},Z) :=& \text{KL}\left( Z \| \text{softmax}(\Tilde{Y}/\tau) \right),  
\end{split}
\end{equation}
where, with a slight abuse of notation, $Z$ is the one-hot distribution of the tokens in the original ending. That is, we encourage the generated logits to recover the original ending.

\paragraph{Ranking}
We rank the candidates based on both their coherence with the context, as well as the internal coherence between the multiple sentences of each candidate (rewritten ending, consists of 3 sentences). More concretely, given a candidate $Y$, we compute the aggregated coherence score:
\begin{equation}
\resizebox{0.89\hsize}{!}{%
$\operatorname{ranking\_score}(Y) = c(X, Y) + \sum_{s=1}^{S-1} c(Y[s], Y[s+1]),$%
}
\label{eq:counter-ranking}
\end{equation}
where each candidate has $S$ sentences (here, $S = 3$) and $Y[s]$ denotes the $s$th sentence. 

\paragraph{Hyperparameters}
We largely follow the same setting as in the abductive reasoning task, but tune hyperparameters on the \dataName development set. Deviations from these settings are outlined in Appendix~\ref{app:task:counterfactual}.

\subsection{Experimental Setup}
\label{sec:tasks:counterfactual:task_setup}

\paragraph{Baselines} 
We compare our method with baselines from \newcite{qin2019counterfactual}. %, which can be categorized into several classes that rely on increasing supervision signal.
The zero-shot baseline uses the pre-trained GPT-2 model to generate $Y$ as a continuation to the counterfactual condition $X$. It is the most apt comparison to our method which also doesn't require additional supervision. We also experiment with two baselines that fine-tune GPT-2 on the original story $X^{ori}Z$ to fit the model to the story domain, either with an LM objective (\texttt{FT}) or a tailored conditional objective that encourages minimal edits of $Z$ (\texttt{Recon+CF}).\footnote{See \newcite{qin2019counterfactual} for more details.} Finally, we report the performance of a supervised baseline (Sup), in which GPT-2 is fine-tuned to produce the gold $Y$ from $X^{ori}Z$ and $X$.

\subsection{Results}
\label{ssec:counterfactual_results}

\paragraph{Automatic Evaluation} Following \newcite{qin2019counterfactual}, we report \textsc{BertScore} \cite{Zhang2019BERTScoreET}, which was shown to best correlate with human judges' notion of counterfactual coherence, and BLEU-4 and ROUGE-L, which better measure minimum-edits. We find that the discriminative baselines achieve the highest degree of plot fidelity. Meanwhile, \modelname{} achieves the highest \textsc{BertScore} for counterfactual coherence.

\paragraph{Human Evaluation}
We repeat the human evaluation setup from Section~\ref{ssec:abductive_results}. Presented with the original story, the counterfactual condition $X$, and the generated ending $Y$, workers were asked to judge (1) the \emph{coherence} of $Y$ with respect to the $X$; and (2) to what extent the generated ending \emph{minimally-edits} the original ending.\footnote{Fair inter-rater agreement with Fleiss' $\kappa = 0.34$} 
In order to judge both criteria, we report the weighted harmonic mean $H_{\beta}$ of these scores across a range of weights $\beta$  (Figure~\ref{fig:cf_curve}).  %\footnote{The average inter-rater agreement measured by Fliess' Kappa \cite{fleiss1971measuring} 
% for two questions is 0.34 (``fair'').}. 

% \begin{equation}
%     H_{\beta} = (1 + \beta^2)\frac{coherence \times similarity}{\beta^2coherence + similarity}
% \end{equation}
% where $coherence$ and $similarity$ are the scores defined above and $\beta$ is a scaling function balancing the importance of both scores.

Our results show that \modelname{} is the only model that maintains a consistent balance between \textit{coherence} (1.66) and \textit{minimal edits} (1.54). While the ranking-augmented zero-shot model produces the most coherent endings (\textit{coherence} = 1.8), it deviates from the original ending. As $\beta$ is increased (i.e., increasing importance of \textit{minimal edits}), its weighted performance drops considerably, indicating it cannot generate new endings that follow the original plot of the story (\textit{min-edit} = 1.25). Conversely, \texttt{Recon+CF} generates stories that are faithful to the original endings, but are far less coherent with the counterfactual condition (\textit{coherence} = 1.23). Through human annotation, we found that \texttt{Recon+CF} copies the original ending word-for-word in a 84\% of cases. 

% Our results show that \modelname{} is best able to balance both coherence and storyline similarity. While the zero-shot model that is augmented with our re-ranker (\S\ref{ssec:framework:ranking}) produces the most coherent endings with respect to the counterfactual sentence ($coherence$ = 1.8), it also struggles to follow the original plot of the story. As $\beta$ is increased (i.e., increasing importance of minimum edits), the mean drops considerably, indicating it cannot generate new endings that follow the original plot of the story ($similarity$ = 1.25). Conversely, the \texttt{Recon+CF} is able to generate stories with a high degree of plot fidelity to the original endings, but rarely produces new endings that are coherent with respect to the counterfactual ($coherence$ = 1.23). In essence, they copy the original ending word-for-word in a large number of cases \antoine{is there a number we can put here to validate this}. \qin{yes, Will add the number.} In contrast, \modelname~is able to balance both coherence ($coherence$ = 1.66) and minimal-edits ($similarity$ = 1.54) to produce counterfactual re-writings with both properties.

% Our results in head-to-head comparisons in Table~\ref{tab:results} parallel these observations as \modelname~significantly outperforms discriminative approaches, \texttt{Recon+CF} and \texttt{Sup+Disc}, in coherence while falling short of the Zero-shot re-ranked baselines. In minimal edits, this pattern is flipped with our approach outperforming Zero-shot baselines considerably and losing to the discriminative baselines.

The pairwise comparison results in Table~\ref{tab:results} parallel these observations. \modelname{} significantly outperforms the discriminative approaches (\texttt{Recon+CF} and \texttt{Sup+Disc}) in coherence, while falling short of the Zero-shot re-ranked baselines. In minimal edits, this pattern is flipped with our approach outperforming Zero-shot baselines considerably and losing to the discriminative baselines. 

\paragraph{Qualitative Analysis}
Figure~\ref{fig:fig_exp_ctf} provides two example results for counterfactual story rewriting by \modelname. The approach successfully captures the causal relations between events and properly rewrites the endings with minimal edits. For instance, in the first example, given the counterfactual condition that ``Tara ordered a shirt online'' (as opposed to the original ``went to mall''), the rewritten ending is about ``sent shirt'' to Tara (as opposed to the original ``browsed from stores''). The last sentence of the original ending ``She looked forward to wearing it'' is correctly preserved as it is coherent with the counterfactual condition.

%% file: emnlp2020-templates/figure/counterfactual_automatic.tex
\begin{table}[t]
\small
    \centering
    \begin{tabular}{lrrr}
    \cmidrule[\heavyrulewidth]{1-4}
      & BLEU\_4 & ROUGE\_L & BERT \\ 
    \cmidrule[\heavyrulewidth]{1-4 }
      \rowcolor[gray]{0.95} \multicolumn{4}{l}{\it Supervised 
       + Discriminative}  \\
    % \rowcolor[gray]{0.95} \multicolumn{2}{l}{\it Supervised + Discriminative} & \multicolumn{2}{r}{\it Input:  $\bm{s}_1\bm{s}_2\bm{y}[S]\bm{s}_1\bm{s}'_2$}  \\
 \textit{Sup+Disc}      & 75.71     & 72.72      & 62.39 \\
    %  \cmidrule{1-6}
     \rowcolor[gray]{0.95} \multicolumn{4}{l}{\it Unsupervised+ Discriminative}  \\
     \textit{Recon+CF}      & \textbf{75.92}     & \textbf{70.93}      & 62.49  \\
     \rowcolor[gray]{0.95} \multicolumn{4}{l}{\it Unsupervised}  \\ 
    \textit{FT}            & 4.06      & 24.09      & 62.55  \\
    % \cmidrule{1-6}
    \textit{FT+CF}        & 4.02     & 24.35      & 62.63  \\
    %  \rowcolor[gray]{0.95} \multicolumn{2}{l}{\it Unsupervised + Discriminative}& \multicolumn{2}{c}{\it Input: $\bm{s}_1\bm{s}_2\bm{y}[S]\bm{s}_1[M]$}    \\ 
   
     \rowcolor[gray]{0.95} \multicolumn{4}{l}{\it Pretrained-only}  \\ 
     \zs$_{\bm{s}_1\bm{s}'_2}$            & 1.74      & 21.41      & 59.31   \\
     \zs$_{\bm{s}_1\bm{s}'_2}$-Ranked            & 2.26      & 25.81     & 60.07  \\
    \textbf{\ourmodel{}}    & 21.35      & 40.73     & \textbf{63.36}  \\
    \midrule
    Human               & 64.93          & 67.64         & 61.87      \\
     \cmidrule{1-4}
    \cmidrule[\heavyrulewidth]{1-4}
    \end{tabular}
    \caption{Automatic evaluation results of counterfactual story rewriting, on the test set of \counterdata{}.} % We highlight the best results of the unsupervised methods.} 
    \label{tab:model_results_counterfactual}
\end{table}

%% file: emnlp2020-templates/figure/counterfactual_human_calibration.tex
\begin{figure}[t]
\centering
\includegraphics[width=0.48\textwidth]{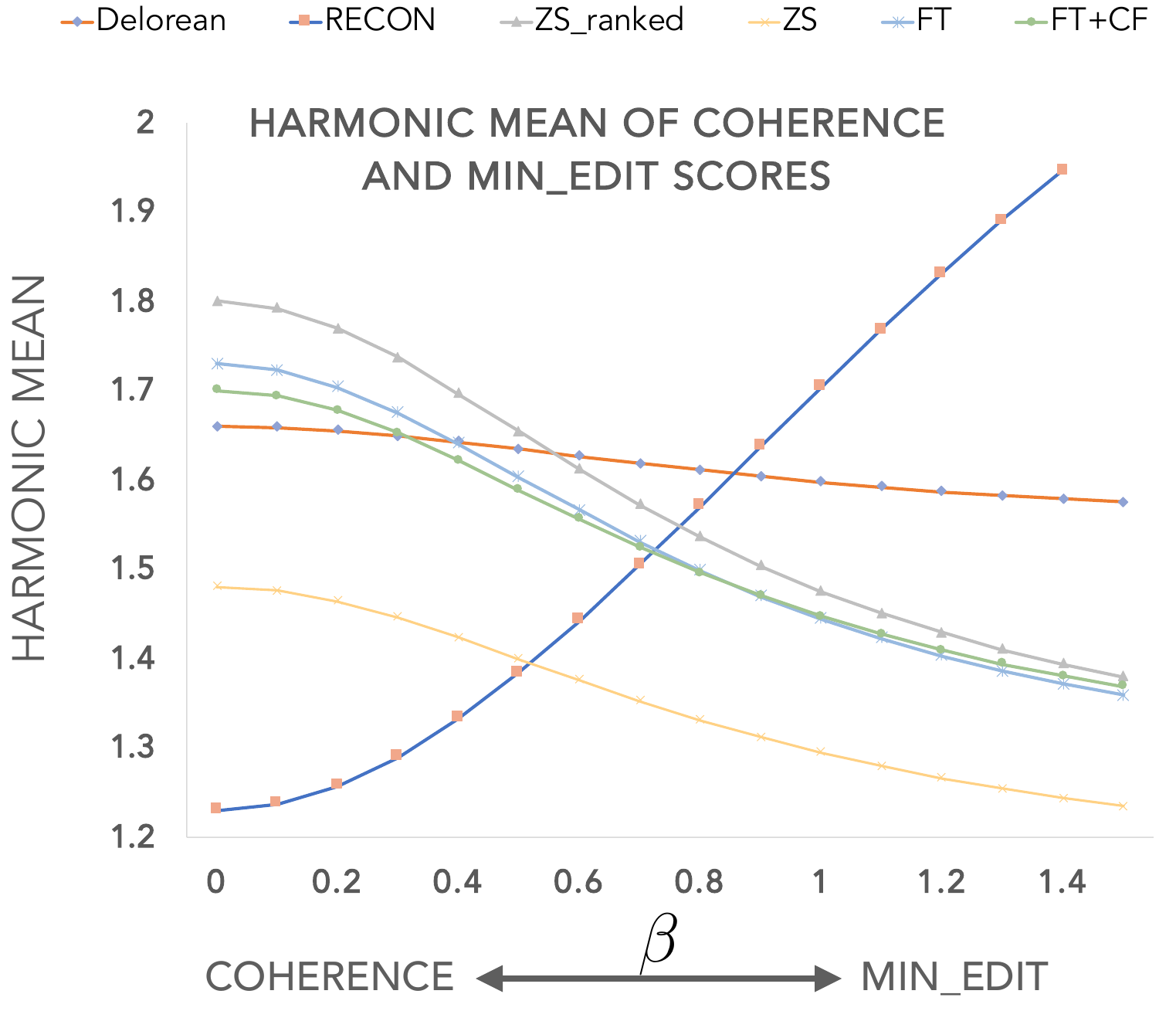}
%\vspace{-15pt}
% \caption{Human calibration results of counterfactual story rewriting. The task has two requirements, namely, coherency and minimal-edits. After ranking, \texttt{ZS}(Zero-shot) works well only on coherency but poorly on minimal-edits. \texttt{Recon+CF} is the best on minimal-edits but the worst on coherency, because its reconstruction loss makes the model usually copy original ending which doesn't fit the counterfactual condition. Our model get a good balance between coherency and minimal-edits.}
\vspace{-15pt}
\caption{
%\qin{update text for 1.5; add $\beta$ to x-axis.}
Human calibration results for counterfactual generation in terms of weighted harmonic mean of coherence and min-edit, $H_{\beta} = \frac{(1 + \beta^2) \cdot \text{coherence} \cdot \text{min\_edit}}{\beta^2 \cdot \text{coherence} + \text{min\_edit}}$, as a function of the scaling factor $\beta$.
Low $\beta$ values assign more weight to coherence, and high $\beta$ values emphasize more on min-edit.
}
\label{fig:cf_curve} 
\end{figure}
\vspace{-10pt}

% After ranking, \texttt{ZS}(Zero-shot) works well only on coherency but poorly on minimal-edits. \texttt{Recon+CF} is the best on minimal-edits but the worst on coherency, because its reconstruction loss makes the model usually copy original ending which doesn't fit the counterfactual condition. Our model get a good balance between coherency and minimal-edits.

%% file: emnlp2020-templates/figure/counterfactual_human_pairwise.tex
\begin{table}[t]
\small
\centering
\scalebox{0.92}{
\begin{tabular}{@{}ll c ll@{}}
\toprule
\multicolumn{5}{c}{Coherence - Human Judges Preferred}                              \\ \midrule
\rowcolor[gray]{0.95} \multicolumn{2}{r}{Our model} & \multicolumn{1}{c}{Neutral} & \multicolumn{2}{l}{Comparator} \\
\ourmodel{} & \textbf{25\%} & 58\%  & 17\%  &  Sup+Disc         \\
\cmidrule{1-5}
\ourmodel{} & \textbf{23\%} & 70\%  & 7\% & Recon+CF    \\
\ourmodel{} & 22\% & 48\%  & \textbf{30\%} & FT            \\
\cmidrule{1-5}
\ourmodel{} & 18\% & 60\% & \textbf{22\%} & \zs$_{\bm{s}_1\bm{s}'_2}$     \\
\ourmodel{}& 27\% & 42\% & \textbf{31\%}  & \zs$_{\bm{s}_1\bm{s}'_2}$-Ranked  \\ 
\cmidrule{1-5}
\ourmodel{} & 10\% & 29\% & \textbf{61\%} & Human   \\
 \bottomrule
\vspace{0.2mm}
\end{tabular}
}
\\
\scalebox{0.92}{
\begin{tabular}{@{}ll c ll@{}}
\toprule
\multicolumn{5}{c}{Min-Edits - Human Judges Preferred}                              \\ \midrule
\rowcolor[gray]{0.95} \multicolumn{2}{r}{Our model} & \multicolumn{1}{c}{Neutral} & \multicolumn{2}{l}{Comparator}  \\
\ourmodel{} &  4\% & 17\%  & \textbf{79\%} & Sup+Disc      \\
\cmidrule{1-5}
\ourmodel{} & 1\% & 14\%  & \textbf{85\%} & Recon+CF    \\
\ourmodel{} & \textbf{21\%} & 76\%  & 3\% & FT            \\
\cmidrule{1-5}
\ourmodel{} & \textbf{28\%} & 71\% & 1\%  & \zs$_{\bm{s}_1\bm{s}'_2}$     \\
\ourmodel{}& \textbf{37\%} & 56\% & 7\%  & \zs$_{\bm{s}_1\bm{s}'_2}$-Ranked  \\ 
\cmidrule{1-5}
M+Sup & 8\% & 22\% & \textbf{70\%} & Human    \\
\bottomrule
%\vspace{0.2mm}
\end{tabular}
}
\vspace{-5pt}
\caption{Human pairwise comparison results on the counterfactual task, between our best model and each baseline with respect to coherence and min-edits.} %``Neutral'' means both are ``equally good/bad''. }
%Percentage of ``equally bad'' is omitted.
\label{tab:results}
\end{table}

%% file: 6-related.tex
\section{Related Work}
\label{sec:related}

\paragraph{Unsupervised text generation.}

Unsupervised approaches are often applied to problems that copy information from a source text into decoded text. Unsupervised paraphrasing requires repeating this information \cite{cgmh, DSSVAE}, as does translation, but with a bilingual transformation \cite{artetxe2017unsupervised, lample2018mt}. In summarization there is an additional task to select a subset of the original text \cite{seq3, schumann2020discrete, bottlesum}. In cases where information is mostly copied from the original, auto-encoding objectives can ensure the correct information is captured \cite{DSSVAE, seq3, artetxe2017unsupervised}. This work tackles problems where generation is more open-ended. Rather than reproducing information from the prompt, generations should agree with and expand on it, making autoencoding less applicable. %\modelname{} solves this by modeling the natural objectives of these problems, with the added benefit that no new model needs to be trained.  \modelname{} solves this by using existing networks that naturally measure the natural objectives of these problems, with the added benefit that no new network needs to be trained. 

\paragraph{Controllable language generation.} Earlier approaches for controllable generation involved preserving the content of text while changing it along discrete dimensions, such as theme, sentiment, or style \cite{koncel2016theme,hu2017toward,ficler2017controlling,shen2017style,lample2019multiple}. Recent works such as Grover \cite{zellers2019defending} and CTRL model \cite{keskar2019ctrl} used these ideas to augment transformer language models that can condition on structured metadata such as source, domain, etc. The Plug \& Play model \cite[PPLM;][]{dathathri2019plug} controls topic and sentiment in an approach similar to ours that involves forward and backward passes to update token distributions. However, PPLM relies on trained attribute discriminators for supervision, while our method is unsupervised. 
%Second, PPLM backpropagates gradients at the token level while we do so at the sequence level. 
While these models are restricted to specific dimensions, often with pre-defined values, our model can adjust to any open-ended textual constraint. 
Perhaps the most similar work in that aspect is the ``text infilling'' models, which, however, are in a more narrow setting by filling only 
%a single or few tokens
a relatively short text span
~\citep{devlin2018bert,zhu2019text,donahue2020infilling}, and more restrictive due to the reliance on an extra right-to-left language model~\citep{sun2017bidirectional} or a pre-specified generation length~\cite[][which is not publicly available]{haim}.

\paragraph{Reasoning about narratives.}

A prominent resource from recent years is the RocStories corpus \cite{Mostafazadeh2016ACA}, consisting of 98K crowdsourced 5-sentence everyday life stories. It was used for the story cloze task whose goal was to predict the story ending from its first 4 sentences, but gained popularity and became the base of additional benchmarks \cite{rashkin2018modeling}. %, and the benchmarks targeted in this paper.  
Additional related work includes ``script knowledge'', i.e. learning about prototypical series of events \cite{schank1977scripts,chambers2008unsupervised,pichotta2014statistical}, temporal commonsense \cite{granroth2016happens,liconstructing}, and modeling pre- and post- conditions of events \cite{roemmele2011choice,Sap2019ATOMICAA,bosselut-etal-2019-comet}. \citet{qin2019conversing} studied conversation modeling that reads and connects the dots of events in related documents.
Finally, a recent line of work explores counterfactual questions in reading comprehension \cite{huang2019cosmos,tandon2019wiqa}, but instantiates the problem of counterfactual reasoning as a multiple choice task.

% Choice of Plausible Alternatives \cite[COPA;][]{roemmele2011choice} is a benchmark for testing causes and effects between simple sentences. 

%% file: 7-conclusion.tex
\section{Conclusion}
\label{sec:conclusion}

% We have presented an unsupervised approach that uses pretrained LMs, such as GPT-2, for abductive and counterfactual reasoning and generation. Our approach interleaves forward and backward passes that iteratively inject both past and future contexts and refine the generation. We further introduced a BERT-based unsupervised ranking method for selecting the best generation from candidates. Our approach performs significantly better than a range of other unsupervised methods. It is interesting to explore the application of the proposed method in other challenging generative reasoning problems.

% We presented \modelname, an unsupervised model that uses pre-trained LMs to generate texts that condition on both a past context and a future constraint, through forward and backward passes considering each condition. We demonstrated the effectiveness of our method for abductive and counterfactual reasoning, on which it performed substantially better than other unsupervised methods. Our proposed method is general and can potentially be used for additional generative reasoning tasks. 

We presented \modelname, an unsupervised LM-based approach to generate text conditioned on past context as well as future constraints, through forward and backward passes considering each condition. We demonstrated its effectiveness for abductive and counterfactual reasoning, on which it performed substantially better than unsupervised baselines. Our method is general and can be easily adapted for other generative reasoning tasks. 

\section*{Acknowledgements}
We thanks the anonymous reviewers and colleages at UW NLP and AI2 for many helpful comments. 
This research was supported in part by DARPA CwC through ARO (W911NF15-1-0543), DARPA MCS program through NIWC Pacific (N66001-19-2-4031), and Allen Institute for AI.

%% file: appendix.tex
\appendix

\section{Additional Experiment Conditions}
\label{app:task}

\subsection{Abductive Reasoning}
\label{app:task:abductive}

%\paragraph{Hyperparameters} 
We set the hypothesis length $N=15$ in the backward pass and allow the forward pass to generate N*2 tokens for complete sentences. We run $T=20$ forward-backward iterations, with each backward pass performing $20$ gradient updates using a small step size $\lambda=0.0003$. The mixing weight of forward/backward logits is $\gamma=0.88$. We use greedy decoding to produce a single candidate at each iteration $T$.

\subsection{Counterfactual Reasoning}
\label{app:task:counterfactual}

%\paragraph{Hyperparameters}  
We use a step size $\lambda=0.0004$ in backward pass and a mixing weight $\gamma=0.92$. One difference from the abductive task is that, here we 
vary the number of forward-backward iterations within $\{5, 10\}$ and the number of backward gradient updates within $\{5, 8, 10, 15\}$. Each configuration produces one candidate at the end of the algorithm. So for each example, we produce 8 candidates for ranking. We found such a generation-ranking protocol gives better performance on the counterfactual task.

Since we need to generate 3 sentences, the number of tokens $N$ is relatively large. For the effectiveness of backpropagation and forward computation, we split the generation into 3 segments, one for each sentence, and perform the forward-backward passes for each segment separately. A sentence that was generated for the $i$th segment, is then appended to the context when generating the $i$+1 segment.